# Exploiting Syntactic Structure for Better Language Modeling: A Syntactic Distance Approach

**Wenyu Du**[1,2][*], **Zhouhan Lin**[3,4][*], **Yikang Shen**[3,4], **Timothy J. O'Donnell**[3,5,6],
**Yoshua Bengio**[3,4] and **Yue Zhang**[1,2]
[1]School of Engineering, Westlake University
[2]Institute of Advanced Technology, Westlake Institute for Advanced Study
[3]Mila    [4]Université de Montréal
[5]Department of Linguistics, McGill University    [6]Canada CIFAR AI Chair, Mila

## Abstract

It is commonly believed that knowledge of syntactic structure should improve language modeling. However, effectively and computationally efficiently incorporating syntactic structure into neural language models has been a challenging topic. In this paper, we make use of a multi-task objective, i.e., the models simultaneously predict words as well as ground truth parse trees in a form called "syntactic distances", where information between these two separate objectives shares the same intermediate representation. Experimental results on the Penn Treebank and Chinese Treebank datasets show that when ground truth parse trees are provided as additional training signals, the model is able to achieve lower perplexity and induce trees with better quality.

## 1 Introduction

It is widely believed in linguistics, cognitive science, and computational linguistics that the latent structure underlying how words combine to form sentences is best represented as a tree structure. The study of the computational mechanisms and systems of constraints that characterize such *derivations* or *parse trees* is a central question in these fields (Pollard and Sag, 1994; Steedman and Baldridge, 2011; Huddleston and Pullum, 2002; Adger, 2003; Bresnan, 2001; Chomsky, 1995; Sag et al., 2003).

Using syntactic information for the language modeling task has been a popular research topic since the 1990s. Early efforts included various approaches that attempted to incorporate shallow syntactic information such as POS tags (Heeman and Allen, 1997; Srinivas, 1996) as well as a more complete structures (Wright et al., 1994; Jurafsky et al., 1995). Most of such work falls under the topic of *structured language modeling* (Chelba and Jelinek, 2000; Van Uytsel et al., 2001; Xu et al., 2002). With the resurgence of neural network approaches, sequential, large-scale neural language models have been shown to significantly outperform traditional language models (Merity et al., 2017; Yang et al., 2018) without using syntactic structural information. On another scenario, recent analysis also reveals that state-of-the-art sequential neural language models still fail to learn certain long-range syntactic dependencies (Kuncoro et al., 2018). Thus it is an interesting problem to explore the relation between language models and syntax and investigate whether syntax can be integrated to enhance neural language models.

To this end, two main lines of work have been investigated, namely transition-based and distance-based methods, respectively. The former strand of work has sought to jointly train a transition-based parser (Nivre, 2008; Zhang and Nivre, 2011; Andor et al., 2016) with a language model using a linearized structured sentence. For example, recurrent neural network grammars (RNNGs) model the joint probability of both words and trees by training a generative, top-down parser (Dyer et al., 2016; Cheng et al., 2017). Subsequent work (Kim et al., 2019b) has developed an unsupervised variant of RNNGs based on an expectation maximization algorithm, which enables the system to be used as a language model without access to parser data.

The second strand of work designs language models that are constrained using syntactic constituents induced using the notion of *syntactic distance* (Shen et al., 2017, 2018). The distances are a sequence of scalars between consecutive words, which are higher when there is a higher level of constituent boundary between the corresponding pair of words. While aligning nicely with the sequential nature of language models, syntactic distances can be transformed into syntactic tree structures with simple principles (Shen et al., 2017).

---
[*]Equal contribution.

The major difference between the above two strands of work is that the former focuses more on parsing performance while the latter aligns better to language model settings. There are three main benefits of the syntactic distance approach. First, typical engineering tricks for language modeling such as batching and regularization (Merity et al., 2017) can be directly used. Second, unlike transition-based methods, which requires to model each sentence independently, distance-based models allow direct comparison with mainstream prior work on language modeling (Gal and Ghahramani, 2016; Merity et al., 2017; Yang et al., 2018) on the same datasets, which carry information across sentence boundaries. Third, there is no risk of compounding errors as compared to the transition-based approach. However, unlike for transition-based approaches (Kim et al., 2019b), for distance-based approaches there have been no studies examining the relationship between induced syntactic structure and human labeled syntactic structure, or whether human labeled syntactic trees can be used to improve language modeling (Dyer et al., 2016; Kim et al., 2019b).

To this end, we investigate distance-based language models with explicit supervision. In particular, we inject syntactic tree supervision into distance-based neural language models by breaking a syntactic tree into a label sequence, and extending a distance-based language model to include a multi-task objective that also learns to predict gold-standard labels. We choose the Ordered-Neuron LSTM (ON-LSTM) (Shen et al., 2018) as our baseline model, which gives the best results among distance-based models.

For making fair comparison with the dominant methods on language modeling, we also manually extend the most commonly-used dataset for evaluating language models, which we name PTB-Concat (Mikolov et al., 2010). It is a version of the Penn Treebank (PTB) (Marcus et al., 1993) dataset with syntactic trees removed, and with preprocessing of numbers, punctuation and singleton words. We add syntactic trees, thus directly compare distance-based methods with other language models.

Experimental results show that incorporating linguistically motivated structures could practically improve language modeling performance. To the best of our knowledge, this is the first work to successfully incorporate gold-standard syntactic trees into syntactic distance based language models. Additional experiments suggest that the level of improvement could also be achieved in other language models. Furthermore, analyses of the trees learned by the multi-task models demonstrate that they are different from both gold trees and unsupervisedly learned trees. [1]

## 2 Related Work

Using syntactic information for language modeling dates back to the last century. Srinivas (1996) proposed using shallow syntactic structures—so-called "super-tags"—which successfully reduced perplexity by 38% over a tri-gram based word-level language model. More complete parser integration is also explored under the heading of "structured language modeling" (Chelba and Jelinek, 2000). This research covers a wide range of different parsers, albeit mostly with N-gram models (Van Uytsel et al., 2001; Xu et al., 2002). Wright et al. (1994) and Jurafsky et al. (1995) extend bi-gram language models with a context-free grammar. Feed-forward neural language models were also explored (Xu et al., 2003). However, the performance does not approach that of the modern neural LMs.

Dyer et al. (2016) first proposed RNNG. Subsequent work extends the model with an encoder-decoder architecture (Cheng et al., 2017), unsupervised learning (Kim et al., 2019b), knowledge-distillation (Kuncoro et al., 2019) and computational psycholinguistics (Hale et al., 2018). Shen et al. (2017) first used syntactic distance to constrain language modeling. Its subsequent work (Shen et al., 2018) transfers the distance notion to LSTM cell. Our work extends distance-based methods in trying to introduce supervised syntax to these models. A very recent work makes use of attention over spans instead of syntactic distance to inject inductive bias to language models (Peng et al., 2019). However, the time complexity of injecting supervision is much higher than distance-based approach ($\mathcal{O}(n^2)$ VS $\mathcal{O}(n)$).

## 3 Model

The overall structure of our model is shown in Figure 1. In particular, the ON-LSTM is taken as the base language model, and syntactic trees are added by conversion to distance metrics. The supervised distance values are taken as one additional output, resulting in a multi-view model.

---

[1] We release the code at https://github.com/wenyudu/SDLM.

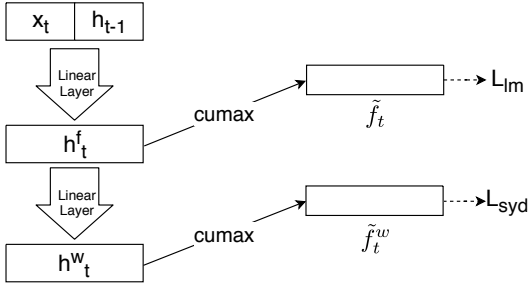

Figure 1: Split-head approach of constructing the two master forget gates in the multi-task setting.

### 3.1 Ordered Neurons LSTM

Ordered Neurons LSTM (ON-LSTM) (Shen et al., 2018) is built upon a vanilla LSTM model (Hochreiter and Schmidhuber, 1997) with two additional gates, namely a master input gate $\tilde{i}_t$ and a master forget gate $\tilde{f}_t$, each being a vector of the same shape as the LSTM forget and input gates:

$$f_t = \sigma(W_f \circ [x_t, h_{t-1}] + b_f) \quad (1)$$
$$i_t = \sigma(W_i \circ [x_t, h_{t-1}] + b_i) \quad (2)$$
$$o_t = \sigma(W_o \circ [x_t, h_{t-1}] + b_o) \quad (3)$$
$$\hat{c}_t = \tanh(W_c \circ [x_t, h_{t-1}] + b_c) \quad (4)$$
$$\tilde{f}_t = \operatorname{cumax}(W_{\tilde{f}} \circ [x_t, h_{t-1}] + b_{\tilde{f}}) \quad (5)$$
$$\tilde{i}_t = 1 - \operatorname{cumax}(W_{\tilde{i}} \circ [x_t, h_{t-1}] + b_{\tilde{i}}) \quad (6)$$

where cumax is defined as the cumulative sum of softmax outputs, i.e., $\operatorname{cumax}(\cdot) = \operatorname{cumsum}(\operatorname{softmax}(\cdot))$. The cumax function provides an inductive bias to model hierarchical structures through enforcing units in the master forget gate $\tilde{f}_t$ to increase monotonically from 0 to 1 and those in the master input gate $\tilde{i}_t$ to decrease monotonically from 1 to 0. The two gates are applied on the original input and forget gates as follows:

$$\omega_t = \tilde{f}_t \circ \tilde{i}_t \quad (7)$$
$$\hat{f}_t = f_t \circ \omega_t + (\tilde{f}_t - \omega_t) = \tilde{f}_t \circ (f_t \circ \tilde{i}_t + 1 - \tilde{i}_t) \quad (8)$$
$$\hat{i}_t = i_t \circ \omega_t + (\tilde{i}_t - \omega_t) = \tilde{i}_t \circ (i_t \circ \tilde{f}_t + 1 - \tilde{f}_t) \quad (9)$$
$$c_t = \hat{f}_t \circ c_{t-1} + \hat{i}_t \circ \hat{c}_t \quad (10)$$
$$h_t = o_t \circ \tanh(c_t). \quad (11)$$

ON-LSTM can learn the implicit structure of a language in the form of a binary tree in an unsupervised manner, through *syntactic distances*, which are calculated as:

$$d_t = D_m - \sum_{k=1}^{D_m} \tilde{f}_t \quad (12)$$

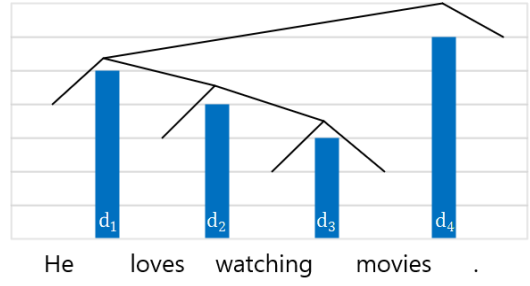

Figure 2: Binarized grammar tree and its corresponding syntactic distances. The heights of the bars stand for the values of the distances. To convert this tree to syntactic distances, we first assign all the words an initial value of 1, and then the non-leaf nodes are assigned distances in the order of $d_3 \to d_2 \to d_1 \to d_4$, according to the procedures in the second part of Model section. On the other hand, given the distances, the tree can be recovered in a top-down process by setting up the split boundaries in descending order of distances (i.e., $d_4 \to d_1 \to d_2 \to d_3$). Syntactically, a shorter distance between a pair of words indicates a closer relationship between the constituents on the two sides of the distance. Note that since only the relative order of the distances could affect the structure of the trees, valid values of these distances are not unique.

where $D_m$ is the size of the hidden state. The syntactic distance $d_t$ between two consecutive words is a scalar value, which can be interpreted as reflecting the syntactic relatedness between the constituents before and after time point $t$. In terms of trees, it can be thought of as the height the lowest tree node that encloses both words. In the case where we consider discrete trees, the height is given by the maximum path length from a leaf. In the more general case, it can be thought of as a scalar value measuring a continuous notion of node height. Figure 2 depicts a sample sentence with its syntactic distances and corresponding tree structures. More generally, the binary tree structure of a sequence with $N$ tokens can be specified with a sequence of $N-1$ syntactic distances. This definition of distance makes the syntactic distance an *ultrametric* (Holly, 2001; Wu et al., 1999), a concept which is important in the theory of hierarchical agglomerative clustering (Johnson, 1967) and was first explored in a linguistic setting by Levelt (1974).

### 3.2 Converting Grammar Trees to Syntactic Distances

To integrate treebank trees into ON-LSTM, we need to first convert syntactic trees into a representation based on syntactic distances. Since the original grammar trees are not necessarily binary, we first split non-binary nodes by adding sentinel intermediate nodes to form a right-branched binary tree, following the steps in Stern et al. (2017). Now for a binary tree with $N$ leaf nodes, we have $N-1$ non-leaf nodes that correspond to the $N-1$ slots between each of the adjacent word pairs, each of which are assigned a syntactic distance (Figure 2). The binary tree can thus be represented as a sequence of distances $d_1, d_2, \ldots, d_{N-1}$.

The conversion from binary tree to syntactic distances thus translates to the assigning of a distance value for each of the $N-1$ non-leaf nodes in the tree. This is achieved in a bottom-up process. We first initialize a distance value of 1 at all of the leaf nodes, and then compute the syntactic distances of the parent nodes by recursively tracing back their parents. More specifically, for a certain parent node, its corresponding syntactic distance $d_P$ is computed with respect to the syntactic distances of its children $d_L$ and $d_R$, i.e.,

$$d_P = \max\{d_L, d_R\} + 1. \qquad (13)$$

A more detailed algorithm flowchart of tree-to-distance conversion is given in Appendix A.1.

### 3.3 Auxiliary Syntactic Distance Outputs

In ON-LSTM the distances $d_t$'s in Equation 12 are used to infer the structure of grammar trees. Consequently, a straight-forward way to incorporate ground truth parse trees is to use the ground truth distances $d_t^g$ to guide $d_t$, as depicted in Figure 1. Interestingly, directly forcing the structure inferred by language models to be coherent to linguist-tagged ground truth trees barely improves the language model performance (see Section 6). Instead, we introduce a "split-head" setting, which can practically improve LM performances by learning two sets of closely related syntactic distances.

In particular, we use another master forget gate $\tilde{f}_t^w$ for inferring a set of distances that are trained to align with the gold-standard syntactic distances, while leaving the original distances $d_t$ computed from $\tilde{f}_t$ intact. To achieve this, we introduce an extra linear layer on top of the hidden states $h_t^f$, and from there infer a separate set of master forget gates. In this way, both of the master forget gates $\tilde{f}_t$ and $\tilde{f}_t^w$ share the same input $h_t^f$, but optimize two different sets of trees for the language modeling and parsing task, respectively. i.e.,

$$h_t^f = W_{\tilde{f}} \circ [x_t, h_{t-1}] + b_{\tilde{f}} \qquad (14)$$
$$\tilde{f}_t = \text{cumax}(h_t^f) \qquad (15)$$
$$\tilde{f}_t^w = \text{cumax}(W_s(h_t^f) + b_s) \qquad (16)$$

The syntactic distances for the auxiliary supervised targets are then calculated as follows:

$$d_t^w = D_m - \sum_{k=1}^{D_m} \tilde{f}_{tk}^w \qquad (17)$$

where $\tilde{f}_{tk}^w$ is the $k$-th element in the vector $\tilde{f}_t^w$

### 3.4 Grammar Trees as Auxiliary Supervised Targets for Language Modeling

With the additional master forget gate $\tilde{f}_t^w$, the model has two different sets of predictions. The first set is the language model outputs of ON-LSTM, predicting the next words. The second set is the distances calculated in Equation 17. The original language modeling structure of the ON-LSTM model is left intact after the modification, so we can continue to use the master forget gate $\tilde{f}_t$ to update hidden states and calculate the softmax output in ON-LSTM for the language modeling part. We denote the negative log-likelihood loss in the language model part as $L_{\text{lm}}$. For brevity, we do not discuss the details of the loss.

For aligning the syntactic distances, we perform a ranking loss between the learned syntactic distance $d_t^w$ and ground truth distance $d^g$, which was first proposed by Burges et al. (2005). The goal is to encourage the model to produce the distances that have the same ranking order as the ground truth distances:

$$L_{\text{syd}} = \sum_{i,j>i} \max(0, (1-\text{sign}(d_i^g - d_j^g)(d_i^w - d_j^w))). \qquad (18)$$

The joint objective function is thus to minimize the following loss:

$$L = L_{\text{lm}} + \alpha L_{\text{syd}} \qquad (19)$$

where $\alpha$ is the scaling parameter.

## 4 Datasets

We make test datasets in English and Chinese, respectively, both of which have parse trees and also language modeling benchmarks. For English, we use the Penn Treebank (PTB) dataset (Marcus et al., 1993). Mikolov et al. (2010) have provided a widely accepted version of PTB for language modeling. Several modifications are made to the original treebank. For example, all punctuation symbols are removed, all characters are lower-cased, the vocabulary size is truncated at 10,000 and all sentences are concatenated. However, this version of PTB discards the parse tree structures, which makes it unsuitable for comparing sequential language models with those utilizing tree structures. We refer to this version as **PTB-Concat**.

Dyer et al. (2016) proposed a different version of PTB, which retains the parse tree structures. Sentences are modeled separately, punctuation is retained, and singleton words are replaced with the Berkeley parser's mapping rules, resulting in much larger vocabulary size, 23,815-word types. Since it retains the parse trees, this dataset enables direct comparison between models that utilize parse trees with those who do not. But unfortunately, since the vocabulary is different from PTB-Concat, and the sentences are processed separately, the results are not directly comparable with those in PTB-Concat, on which most existing work on language modeling reports results. We refer to this version as **PTB-Sepsent**.

As mentioned above, a salient limitation of PTB-Sepsent is that it does not allow fair comparison with existing LM work on PTB-Concat. To address this issue, we propose a different variation of PTB dataset that both uses the same vocabulary size as PTB-Concat and at the same time retaining the ground-truth grammar trees. We pre-process the PTB dataset by following the same steps indicated by Mikolov et al. (2010) to obtain a modified treebank with the same vocabulary set as PTB-Concat. Sentences are concatenated, and we make sure that the sentences are the same to PTB-Concat, from token to token, in the training, validation, and test sets. This results in the same vocabulary as that of PTB-Concat, which allows us to directly compare models that utilize parse trees with the existing reports of performance on PTB-Concat. We refer to this version of PTB-Concat with syntax as **PTB-Concat-Syn** and we will cover preprocessing details in Appendix A.3.

For Chinese, we use the Chinese Treebank 5.1 (Xue et al., 2005), with the same settings as Kim et al. (2019b). Sentences are modeled separately and singleton words are replaced with a single <UNK> token. It will be referred to as **CTB-Sepsent** in the rest of the paper.

| Model | Param | Dev | Test |
|---|---|---|---|
| Gal and Ghahramani (2016) - Variational LSTM | 66M | – | 73.4 |
| Kim et al. (2016) - CharCNN | 19M | – | 78.9 |
| Merity et al. (2016) - Pointer Sentinel-LSTM | 21M | 72.4 | 70.9 |
| Grave et al. (2016) - LSTM | – | – | 82.3 |
| Zoph and Le (2016) - NAS Cell | 54M | – | 62.4 |
| Zilly et al. (2017) - Variational RHN | 23M | 67.9 | 65.4 |
| Shen et al. (2017) - PRPN | – | – | 62.0 |
| Merity et al. (2017) - 3-layer AWD-LSTM | 24M | 60.0 | 57.3 |
| Zolna et al. (2018) - Fraternal dropout | 24M | 58.9 | 56.8 |
| Shen et al. (2018) - 3-layer ON-LSTM | 25M | 58.3 | 56.2 |
| ONLSTM-SYD | 25M | **57.8** | **55.7** |
| Yang et al. (2018) - AWD-LSTM-MoS | 22M | 56.5 | 54.4 |
| Takase et al. (2018) - AWD-LSTM-DOC | 23M | 54.1 | 52.4 |

Table 1: Various language models evaluated on validation and test sets on PTB-Concat. Our model is denoted as ONLSTM-SYD, which incorporates tree structures during training. Yang et al. (2018) and Takase et al. (2018) focus on improving the softmax module of LSTM LM, which are orthogonal to ours.

## 5 Experiments

We evaluate the influence of syntactic supervision on distance-based language models, especially in terms of its language modeling performance. We are also going to analyze the induced syntax after introducing the structural supervision. In addition, extensive ablation tests are conducted to understand how syntactic supervision affects the language model.

### 5.1 Language Modeling

We first compare our models with existing sequential language models on PTB-Concat, and then we compare our model with transition-based language models on PTB-Sepsent and CTB-Sepsent, which have a larger vocabulary and also use additional grammatical structure.

**Results on PTB-Concat** We first validate the benefit of introducing structural signal to neural language models by training our proposed model on PTB-Concat-Syn with structural supervision, and then evaluate them on the plain validation/test set. We compare our model with the original ON-LSTM model, as well as various other strong LSTM language model baselines such as AWD-LSTM (Merity et al., 2017) and a mixture of softmax (Yang et al., 2018). We denote our syntactic-distance-augmented ON-LSTM model as ONLSTM-SYD.

For making fair comparison, we closely follow the hyperparameters and regularization of ON-LSTM (Shen et al., 2018). The model is a three-layer ONLSTM-SYD language model with an embedding size of 400 and hidden layer units 1150.

| Model | PTB-Sepsent | CTB-Sepsent |
|---|---|---|
| Kim et al. (2019b) - RNNLM | 93.2 | 201.3 |
| Kim et al. (2019b) - RNNG | 88.7 | 193.1 |
| Kim et al. (2019b) - URNNG | 90.6 | 195.7 |
| Kim et al. (2019b) - RNNG-URNNG | 85.9 | 181.1 |
| Kim et al. (2019b) - PRPN (default) | 126.2 | 290.9 |
| Kim et al. (2019b) - PRPN (finetuned) | 96.7 | 216.0 |
| ONLSTM-noAWD | 69.0 | 167.7 |
| ONLSTM | 60.0 | 145.7 |
| ONLSTM-SYD-noAWD | 67.6 | 163.1 |
| ONLSTM-SYD | **59.6** | **140.5** |

Table 2: Language modeling perplexity on PTB-Sepsent and CTB-Sepsent. Kim et al. (2019b) report two results of PRPN, the default one using settings in Shen et al. (2017) and another one finetuned by themselves. Our models use the same hyperparameter settings as in Section 5.1.

The dropout rates are $0.5, 0.45, 0.3, 0.45$ for the word vectors, LSTM weight metrics, outputs between LSTM layers and the output of the last layer, respectively. The embedding dropout ratio is $0.125$. The model is trained and finetuned for $1000$ epochs in total and is switched to the fine-tuning phase at epoch $650$. The ground truth syntactic structures are used to supervise the syntactic distances in the third layer of ONLSTM-SYD and the loss raio $\alpha$ is set to $0.75$. We use this setting as the default setting for all the experiments.

The results are shown in Table 1. After adding structural signals into the model, our model ONLSTM-SYD significantly outperforms the original ON-LSTM model ($p$-value $< 0.05$), indicating that incorporating linguist-tagged parse trees can contribute to language modeling positively.

**Results on PTB-Sepsent and CTB-Sepsent** PTB-Sepsent and CTB-Sepsent offer a comparable setting with other structure-aware supervised (Dyer et al., 2016) and unsupervised (Kim et al., 2019b) baselines. The results are listed in Table 2. ONLSTM-SYD performs better than ONLSTM, which indicates that supervised syntactic information can help improve language modeling.

The margin between our models and the baselines is rather large. We find that the set of regularization and optimization techniques proposed by Merity et al. (2017) contribute significantly to this margin. Because of the sequential and parallel nature of our model, it can directly inherit and benefit from this set of tricks. In contrast, it is non-trivial to use them for RNNG and URNNG. As a more rigorous analysis, we further conducted a set of experiments without those tricks (i.e. non-monotonically triggered ASGD, weight-dropped LSTM, finetuning). The performance (denoted as ONLSTM-SYD-noAWD) drops; however, the model still outperforms the other baselines by a significant margin.

### 5.2 Structure Analysis

In this subsection we analyze the model to see how the additional structural supervision affects the quality of inferred trees. Note that our goal here is to analyze the influence of ground truth syntactic information on the quality of the induced trees rather than to yield a better grammar induction performance, since our model is not strictly comparable to other models due to its extra structural supervision during training.

We follow the settings of Htut et al. (2018) to test our model on the WSJ10 and WSJ test sets, reporting the results in Table 3. The WSJ test set has 2416 sentences with arbitrary lengths, while WSJ10 consists of 7422 sentences of the whole WSJ corpora that contain no more than 10 words. We use both biased and unbiased distance-to-tree conversion algorithms for both ON-LSTM and our proposed model (c.f. Appendix A.1 and A.2 for a formal description of the biased and non-biased conversion algorithm). Since our model has two sets of trees learned simultaneously, we list all of them in Table 3.

**Grammar Induction** We can see that the trees learned by the joint loss show improved the F1 score and rely less on the branching bias of the tree constructing algorithm (see Dyer et al. (2019)). The big gap of F1 scores on WSJ between the biased and unbiased trees are altered after introducing the structural loss, and the LM unbiased trees significantly outperforms its baseline ON-LSTM. These indicate that the auxiliary supervised task not only lowers the perplexity, but also improves the qualities of the induced trees for the LM task.

Looking more into the trees, we find that compared to ON-LSTM, ONLSTM-SYD improves the label prediction accuracy for NP (noun phrases), VP (verb phrases) and PP (prepositional phrases) but fails to improve ADJP (adjective phrases). This suggests that different types of human-annotated constituents may have different influences on language modeling, or that human-annotated trees are themselves biased to differing degrees between different constituent types.

**Branching Bias** Syntactic trees of English naturally have a bias towards right branching struc-tures. As shown in the last section of Table 3, right branching trees achieve a much higher F1 score than random, balanced or left branching trees. As pointed out by Dyer et al. (2019), PRPN and ON-LSTM resort to a distance-to-tree algorithm with right-branching biases (See Appendix A.2).

| Model | Training Objective | Induction Algorithm | Parsing F1 WSJ10 | Parsing F1 WSJ | Depth WSJ | Accuracy on WSJ by Tag ADJP | NP | VP | PP | R/L Ratio on WSJ |
|---|---|---|---|---|---|---|---|---|---|---|
| ON-LSTM | LM | Unbiased | 63.2 | 39.0 | 4.9 | 37.9 | 42.8 | 49.6 | **54.2** | 1.08 |
| ON-LSTM | LM | Biased | **69.5** | **44.2** | 5.5 | **57.0** | **53.0** | 52.4 | 49.6 | 2.09 |
| ONLSTM-SYD$^{syd}$ | LM+SYD | Unbiased | **77.6** | **61.3** | 7.3 | 38.2 | **73.2** | 69.6 | **72.9** | 2.81 |
| ONLSTM-SYD$^{syd}$ | LM+SYD | Biased | 65.7 | 45.5 | 5.5 | 30.4 | 40.6 | **70.7** | 43.9 | 5.07 |
| ONLSTM-SYD$^{lm}$ | LM+SYD | Unbiased | 55.1 | 34.5 | 4.8 | 14.9 | 42.2 | 16.7 | **67.4** | 0.83 |
| ONLSTM-SYD$^{lm}$ | LM+SYD | Biased | 58.0 | 36.3 | 5.3 | **41.1** | **53.9** | **52.4** | 43.0 | 1.70 |
| Binary Gold Standard Trees | – | – | 88.1 | 85.6 | 6.4 | 100 | 100 | 100 | 100 | 2.92 |
| Gold standard Trees | – | – | 100 | 100 | 5.0 | 100 | 100 | 100 | 100 | 2.22 |
| Random Trees (Htut et al., 2018) | – | – | 32.2 | 18.6 | 5.3 | 17.4 | 22.3 | – | 16.0 | – |
| Balanced Trees (Htut et al., 2018) | – | – | 43.4 | 24.5 | 4.6 | 22.1 | 20.2 | – | 9.3 | – |
| Left Branching Trees | – | – | 19.6 | 9.0 | 12.4 | – | – | – | – | – |
| Right Branching Trees | – | – | 56.6 | 39.8 | 12.4 | – | – | – | – | – |

Table 3: Unlabeled parsing results evaluated on the WSJ10 and the full WSJ test set. Numbers in bold font indicate that they are the best compared to those computed from the other parts of the model (i.e., within the same section in the table). The *Algorithm* column represents whether bias or unbiased algorithm is performed. ONLSTM-SYD$^{syd}$ and ONLSTM-SYD$^{lm}$ represent two sets of trees induced from loss $L_{syd}$ and $L_{lm}$ respectively. The *Accuracy* columns represent the fraction of ground truth constituents of a given type that correspond to constituents in the model parses. The *R/L Ratio* column represents the ratio between the number of words that are left children of its parent, and those that are right children.

For our model, a biased distance-to-tree algorithm yields worse results compared to its non-biased counterpart; but on unsupervised models such as ON-LSTM, biased algorithms yield better results than non-biased versions. This observation indicates that syntactic supervision leads to better tree structures as compared with fully unsupervised tree induction, which is intuitive.

**Linguistic Analysis** Our best parsing results are for trees decoded from the syntactic prediction objective using the unbiased algorithm. Interestingly, these trees tend to be deeper on average than the (binarized) gold standard trees (see Table 3).[2] This appears to be driven by a failure of the model to identify constituents centered on deeply-embedded head words—instead, the model prefers right-branching structures. Some examples of trees are displayed in Figure 3. In the top part of the figure, we see the parse produced from the $L_{syd}$ distances of our model, in the middle the tree produced the $L_{lm}$ distances and, on the bottom, the gold standard tree. As can be seen in the figure, the $L_{syd}$-based tree is largely right-branching and misses constituents centered on several deeply embedded heads, such as the verb *said*. By contrast, the $L_{lm}$-based tree is considerably shallower than the gold-standard and consists of a sequence of smaller chunks that often mis-bracket words with respect to the gold-standard constituent boundaries.

Figure 4 illustrates these phenomenon in further detail. The plot at the top of the figure shows the proportion of constituents produced from $L_{syd}$ distances whose boundaries correspond to a gold constituent, broken down by height of nodes in the predicted tree. As the plot illustrates, the model fares better on relatively small constituents lower in trees, and makes more errors for constituents higher in the tree, reflecting mistakes on deeply-embedded heads. The bottom of the figure shows the same breakdown for $L_{lm}$-based induced trees. Overall, the affect is similar, although $L_{lm}$-based trees are shallower than the $L_{syd}$-based trees. We believe the increased accuracy for the longest constituents is driven by the fact that, since the highest constituents cover long sentence spans and there are few possible long spans, these constituents have a higher baseline probability of being correct.

It appears that the $L_{syd}$ objective has learned a strong right-branching bias, leading to very deep trees (even with the unbiased decoder) whereas the $L_{lm}$ objective appears to be using a kind of predictive chunking of the sentence into small groups of words. It is tempting to speculate that these chunks may correspond to linguistic units used in prosodic planning or by the human sentence processor, while the deeper trees correspond more directly to the compositional structure underlying sentence meaning. We leave exploring this question to future work.

---
[2]Please refer to Appendix A.5 for visualizations of a more extensive set of sentences.

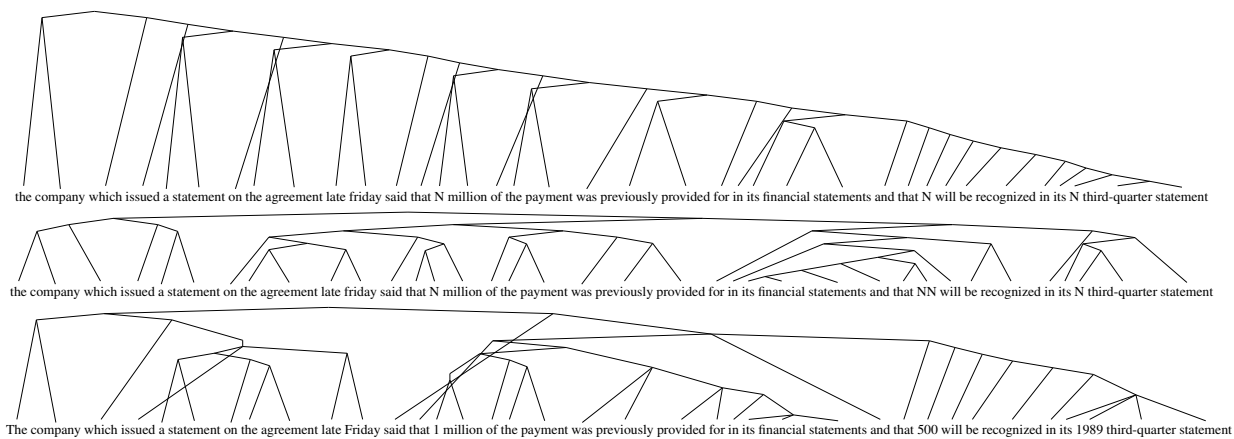

Figure 3: Trees induced from the syntactic task distances in our model (top), the language modeling task distances (middle) as well as the gold-standard trees (bottom).

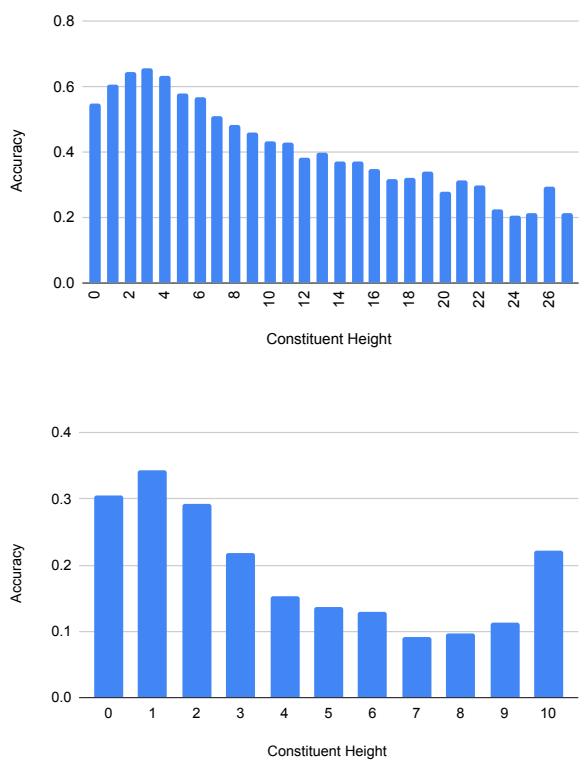

Figure 4: Accuracy breakdown w.r.t. constituent height in unbiased trees derived from the syntactic task distances in our model (top) and the language modeling distances (bottom). A constituent is considered as correct if its boundaries correspond to a true constituent. The constituents' heights are those in the predicted tree. Since constituents that represent the whole sentence always have correct boundaries, they are excluded from the calculation.

**Parsing performance** Our models give worse unlabeled parsing performance compared to transition-based methods. In particular, Kim et al. (2019a) report that unsupervised URNNG achieves 45.4 WSJ F1 in a similar setting, while another URNNG that finetunes a supervised RNNG model gives a much better F1 of 72.8, leading a 27.4 F1 improvement. In contrast, the F1 of our structure prediction trees is 61.3 in unbiased algorithm. This indicates that our model brings more benefits on the LM side rather than the parsing side.

## 6 Ablation Study

**Layer used for supervision** Table 4 (Top) shows the performances where the supervised signal is injected into different layers. Although injecting syntax into the last layer gives the best syntactic distance for grammar induction, it fails to achieve a similar improvement on perplexity. This suggests that a better syntactic structure may not always lead to a better language model. The observation is consistent with prior research (Williams et al., 2018).

**Tree structure** We study the influence of the different types of supervised trees to the model. In addition to using the ground truth parse trees, we also tried to train the model with random trees instead, and without providing trees, in which case it degenerates to a vanilla ON-LSTM. From Table 4 (Middle) we can find that without supervision signals from gold standard parse trees the model performs worse than the full model. Random trees introduce noise to the model and downgrade both parsing and LM performance, indicating the importance of injecting meaningful syntax.

**Multitask variants** We also explored injecting the supervised syntactic information at different levels. One straight forward baseline is to add su-pervision signals directly on the syntactic distance in ON-LSTM, using one set of trees to guide both LM and parsing, as indicated in the Model section (Table 4 Bottom, one set of trees). Despite injecting stronger syntactic signals, this direct approach does not improve language model perplexity. This also reflects the fact that the most suitable syntactic structures for language modeling do not necessarily conform to human labeled syntax. In addition, we also use ON-LSTM hidden states for supervised syntactic distance prediction (Table 4 Bottom, vanilla multitasking). This approach fails to outperform its ON-LSTM baseline due to the same reason. In summary, there are mutual benefits between induced and supervised syntactic information, although they do not fully overlap.

| Ablation Study | Experiment Detail | Validation PPL | Test PPL | WSJ F1 |
|---|---|---|---|---|
| Layer for Supervision | 1st layer | 58.0 | 55.6 | 57.7 |
| | 2nd layer | **57.8** | **55.5** | 59.7 |
| | 3rd layer | **57.8** | 55.7 | **61.3** |
| Tree Structure | No Parse Tree | 58.3 | 55.9 | 39.0 |
| | Random Tree | 60.2 | 57.5 | 32.4 |
| | Gold Parse Tree | 57.8 | 55.7 | 61.3 |
| Multitask Variants | Vanilla Multitasking | 60.9 | 58.5 | 24.9 |
| | One set of trees | 58.5 | 55.9 | 54.4 |
| | Two sets of trees | 57.8 | 55.7 | 61.3 |

Table 4: Perplexity and unlabeled parsing F1 in ablation studies. We choose unbiased algorithm and the layer with supervision injected. For the unsupervised models, we report the layer with best F1 score. (Top) When supervising on different layers. (Middle) Using different tree structures for supervision. (Bottom) Different multitasking strategies.

**Generalization to other LMs** One practical question is whether the improvements found in our work can be generalized to other language models. To answer this question, we introduce the multi-task scheme to PRPN (Shen et al., 2017), which is another model that is also able to learn unsupervised structures through language modeling. Similar to ON-LSTM, PRPN is also a syntactic distance method. We modify the PRPN model in the same spirit as in ON-LSTM. In addition, we change the encoding layer and use the output as syntactic distance embeddings $l_{syd}$. Then we map $l_{syd}$ to two sets of syntactic distances $d_{lm}$ and $d_{syd}$ for language modeling and syntactic distance prediction, respectively. Syntactic supervision comes to $d_{syd}$. The model reaches a test perplexity of 60.5 in PTB-Concat ($p$-value $< 0.05$), which also significantly outperforms the 62.0 from the original model. We refer readers to Appendix A.4 for the details of PRPN and our modified PRPN-SYD.

# 7 Conclusion

We investigated linguistic supervision for distance-based structure-aware language models, showing its strengths over transition-based counterparts in language modeling. Apart from the explicit observations in achieving strong perplexity scores, our model reveals several interesting aspects of the quality of the trees learned by the model. As a byproduct of our investigation, we release a version of PTB-Concat, which contains syntactic structures while at the same time the same pre-processing steps adopted by most previous work on neural language models.

## Acknowledgments

We thank Zhiyang Teng, Qi He and all members at Text Intelligent Lab in Westlake University for insightful discussions. We also would like to thank all anonymous reviewers for their constructive comments. This work is supported by the National Natural Science Foundation of China (NSFC No. 61976180) and the Westlake University and Bright Dream Joint Institute for Intelligent Robotics. The corresponding author is Yue Zhang.

## A Appendices

### A.1 Algorithms for transformation between parse trees and syntactic distances

The following tree-to-distance algorithm provides a set of distances given a tree. The node indicates the root node of the given tree.

**Algorithm 1** Binary Parse Tree to Distance
($\cup$ represents the concatenation operator of lists)
1: **function** TREE2DISTANCE(node)
2:     **if** node is leaf **then**
3:         $d \leftarrow 1$
4:     **else**
5:         $\text{child}_l, \text{child}_r \leftarrow$ children of node
6:         $\mathbf{t2d}_l \leftarrow \text{Tree2Distance}(\text{child}_l)$
7:         $\mathbf{t2d}_r \leftarrow \text{Tree2Distance}(\text{child}_r)$
8:         $d \leftarrow \max(d_l, d_r) + 1$
9:         $\mathbf{t2d} \leftarrow \mathbf{t2d}_l \cup [d] \cup \mathbf{t2d}_r$
10:     **end if**
11:     **return** $\mathbf{t2d}, d$
12: **end function**

The following distance-to-tree conversion algorithm provides an unbiased reconstruction of tree given a set of distances.

**Algorithm 2** Distance to Binary Parse Tree
1: **function** DISTANCE2TREE(**d**)
2:     **if** $\mathbf{d} \neq []$ **then**
3:         $i \leftarrow \arg\max_i(\mathbf{d})$
4:         $\text{child}_l \leftarrow \text{Distance2Tree}(\mathbf{d}_{<i})$
5:         $\text{child}_r \leftarrow \text{Distance2Tree}(\mathbf{d}_{\geq i})$
6:         $\text{node} \leftarrow \text{Node}(\text{child}_l, \text{child}_r)$
7:     **end if**
8:     **return** node
9: **end function**

### A.2 Distance-to-tree algorithm with right-branching bias

**Algorithm 3** Distance to Binary Parse Tree with Right-Branching Bias
1: **function** DISTANCE2TREE(**d**)
2:     **if** $\mathbf{d} \neq []$ **then**
3:         $i \leftarrow \arg\max_i(\mathbf{d})$
4:         $\text{child}_l \leftarrow \text{Distance2Tree}(\mathbf{d}_{<i})$
5:         $\text{child}_r \leftarrow \text{Distance2Tree}(\mathbf{d}_{>i})$
6:         $\text{node}_{bias} \leftarrow \text{Node}(\text{node}_i, \text{child}_r)$
7:         $\text{node} \leftarrow \text{Node}(\text{child}_l, \text{node}_{bias})$
8:     **end if**
9:     **return** node
10: **end function**

### A.3 Details of generating our PTB-Concat-Syn version

Mikolov et al. (2010) briefly described the steps of converting from the original Penn Treebank dataset to his version of dataset, which later becomes the standard in language modeling task. We denote this version as PTB-Concat. In our paper, to get strictly the same PTB language modeling dataset, we follow his steps on the original Penn Treebank, while preserving the tree structure. Specifically, we took the following steps:

    1. Convert all tokens to lowercase.

    2. For tokens which are purely digits, or digits only with "`.`" or "`-`" are converted to token "`N`".

    3. Replace all "`$`" with "`N`".

    4. Delete tokens "`\`" and "`wa`" if their POS tags are "`POS`" and "`NNP`", respectively.

    5. Delete all tokens that fall into the following list:
[``,\',\',,,,.,:,;,-,?,!,¡¿-,\\,|,~,-lrb-,-rrb-,-lcb-,-rcb-,(,),[,],{,},<,>,--,...,`].

    6. Delete all tokens with tag "`-NONE-`".

    7. Add a special token "`</s>`" to the end of each sentence.

    8. Truncated the vocabulary at $9,999$ according to the frequencies and assign all the out-of-vocabulary tokens a special token "`<unk>`".

    9. After the above procedures, there are still minor differences to PTB-Concat. We then go through the whole Penn Treebank corpora to manually fix all the unmatched tokens.

These procedures ensures we have exactly the same training, validation and test sets as PTB-Concat, the only difference is that our datasets has additional grammar trees retained from the original PTB dataset. The resulting datasets then becomes PTB-Concat-Syn.

## A.4 PRPN and PRPN-SYD

### A.4.1 Parse-Read-Predict Network (PRPN)

The idea of PRPN builds upon an assumption that to predict a word $x_i$, we only need information for all precedent siblings in constituent tree. The model constitutes three components: (i) *a parsing network* that calculates the syntactic distance and parsing gates. (ii) *a reading network* to model the language, and (iii) *a predict network* to predict the next word.

PRPN first uses a two-layer convolutional network to calculate the syntactic distance $d$ at timestep $t$:

$$h_i = \text{ReLU}(W_c \begin{bmatrix} e_{i-L} \\ e_{i-L+1} \\ ... \\ e_i \end{bmatrix} + b_c) \quad (20)$$

$$d_i = \text{ReLU}(W_d h_i + b_d) \quad (21)$$

Where $e_{i-L}, ..., e_i$ are word embeddings, $L$ is the lookback range.

Then the difference between distances is fed through hardtanh to model the degree $\alpha_j^t$ that how much two words $x_t$ and $x_j$ are related:

$$\alpha_j^t = \frac{\text{hardtanh}((d_t - d_j) \cdot \tau) + 1}{2} \quad (22)$$

Where $\text{hardtanh}(x) = \max(-1, \min(1, x))$, and $\tau$ is the temperature parameter.

For word $x_i$, the first precedent word $x_t$ with a small value $\alpha_i^t$ represents $x_t$ and all its precedents are not likely to be siblings of $x_i$. The following parsing gate $g_i^t$ models the probability of $x_t$ and $x_i$ being siblings:

$$g_i^t = \mathbf{P}(l_t \leq i) = \prod_{j=i+1}^{t-1} \alpha_j^t \quad (23)$$

The reading network is a variant of Long Short-Term Memory-Network (LSTMN) (Cheng et al., 2016) where the attention score is softly truncated by parsing gates:

$$s_i^t = \frac{g_i^t \tilde{s}_i^t}{\sum_i g_i^t} \quad (24)$$

The predict network utilizes the structure-aware hidden states of reading network to predict the next word.

### A.4.2 The PRPN-SYD model

We re-designed the parsing network. We use LSTM to encode each embedding sequence $s = (e_0, e_1, ..., e_n)$,. Because the task of language modeling prohibits seeing future words, we use unidirectional LSTM:

$$h_0, ..., h_n = \text{LSTM}_w(e_0, ..., e_n) \quad (25)$$

We stack a convolutional layer on top of the hidden states $h_i$ of the LSTM, which helps gather local syntactic information:

$$g_0, ..., g_n = \text{CONV}(h_0, ..., h_n) \quad (26)$$

Next, syntactical information learned both locally and globally are integrated by using another unidirectional LSTM:

$$\hat{h}_0, ..., \hat{h}_n = \text{LSTM}_d(g_0, ..., g_n) \quad (27)$$

We pass the $\hat{h}$ layer through two 2-layer fully-connected networks which output two respective sets of distance scalars:

$$d_i^{lm} = FF_{lm}(\hat{h}_i) \qquad d_i^{syd} = FF_{syd}(\hat{h}_i) \quad (28)$$

Where $d^{lm}$ is the distance for language modeling while $d^{syd}$ is for syntactic distance prediction. For two sets of distances, we use the same objective functions as described in ONLSTM-SYD.

## A.5 Trees

We visualize a set of sentences (14 sentences in total) and their corresponding trees in parallel to contrast the qualitative differences of the model induces trees and gold standard trees. Sentences are selected randomly from the dataset. In each of the following figures, we provide three trees for a same sentence, which corresponds to trees induced from the syntactic task (top) and language model task (middle) set of distances, as well as the gold-standard trees (bottom).

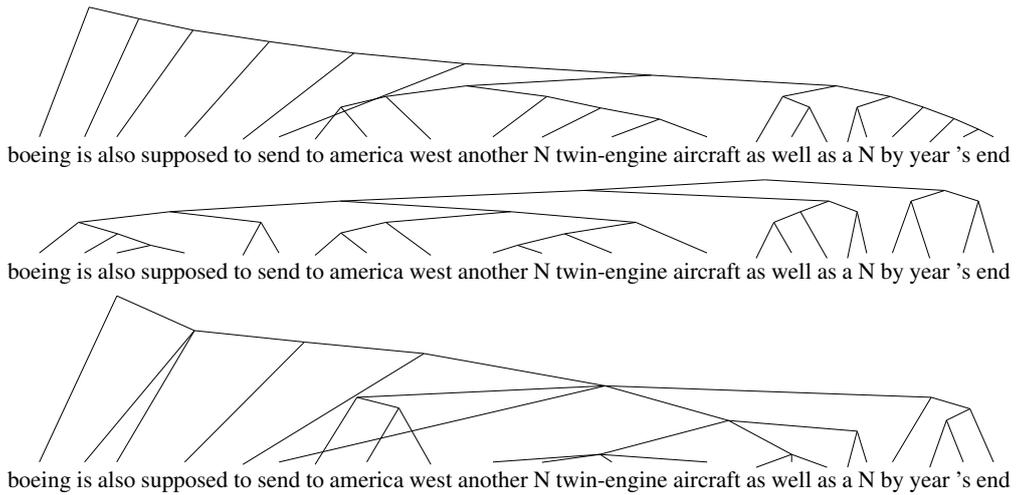

Figure 5: Sentence 1. Trees induced from the syntactic task (top) and language model task (middle) set of distances, as well as the gold-standard trees (bottom).

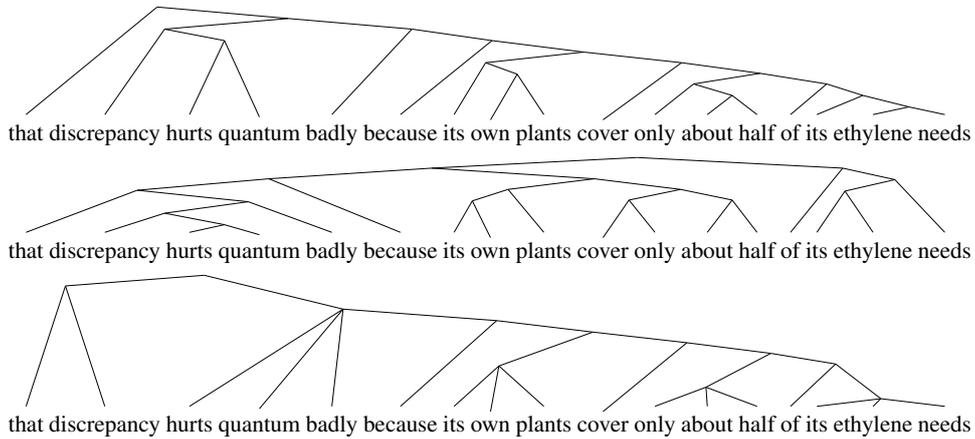

Figure 6: Sentence 2. Trees induced from the syntactic task (top) and language model task (middle) set of distances, as well as the gold-standard trees (bottom).

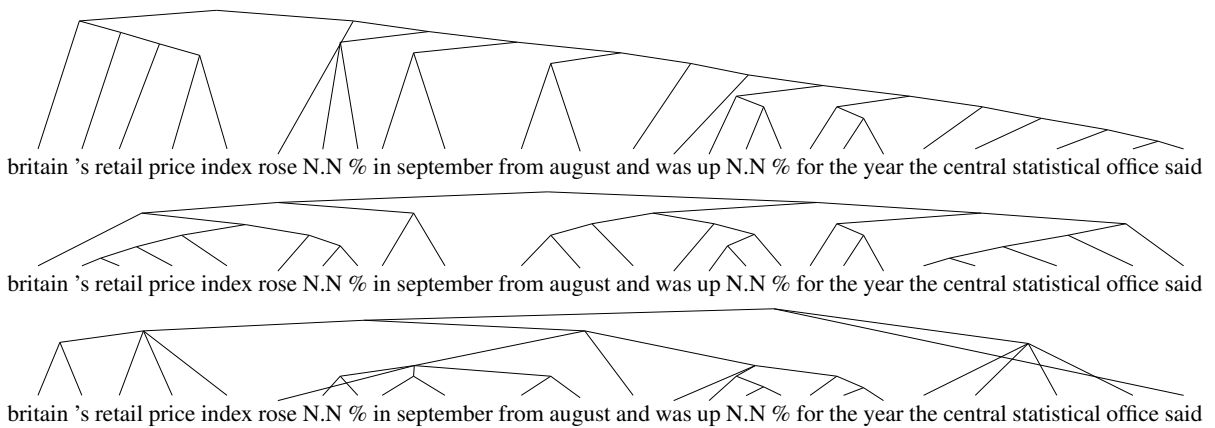

Figure 7: Sentence 3. Trees induced from the syntactic task (top) and language model task (middle) set of distances, as well as the gold-standard trees (bottom).

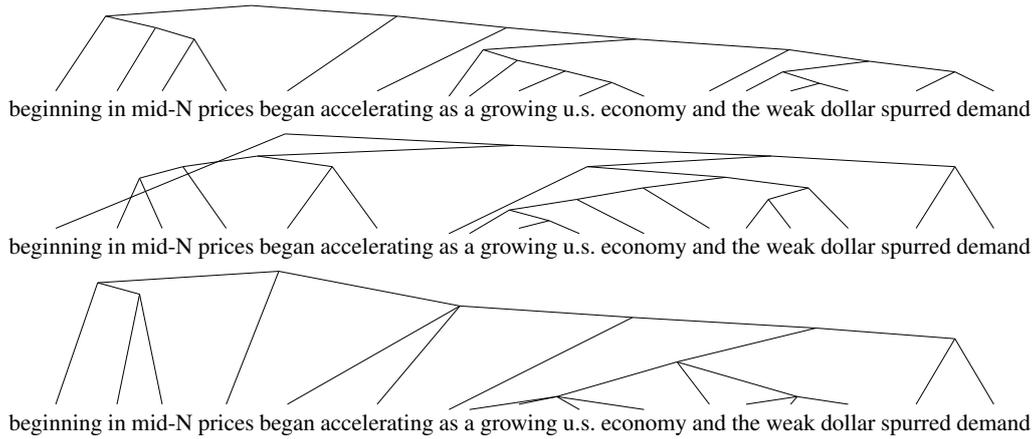

Figure 8: Sentence 4. Trees induced from the syntactic task (top) and language model task (middle) set of distances, as well as the gold-standard trees (bottom).

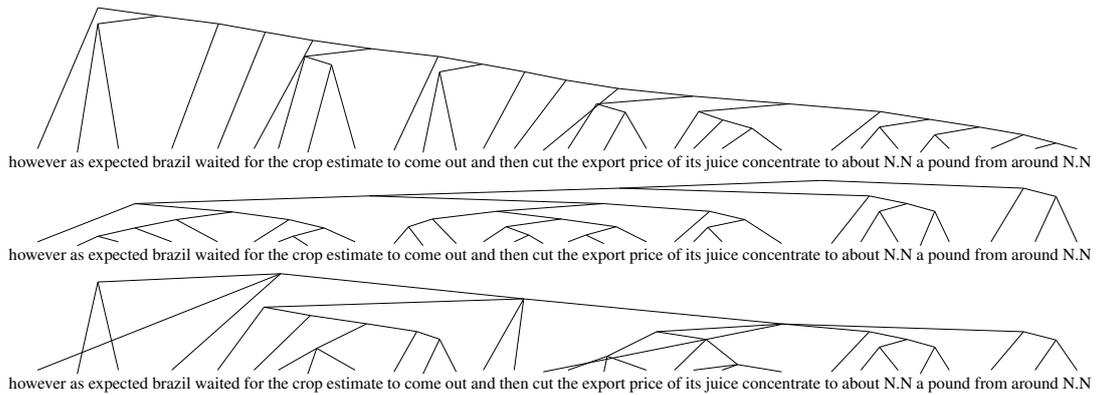

Figure 9: Sentence 5. Trees induced from the syntactic task (top) and language model task (middle) set of distances, as well as the gold-standard trees (bottom).

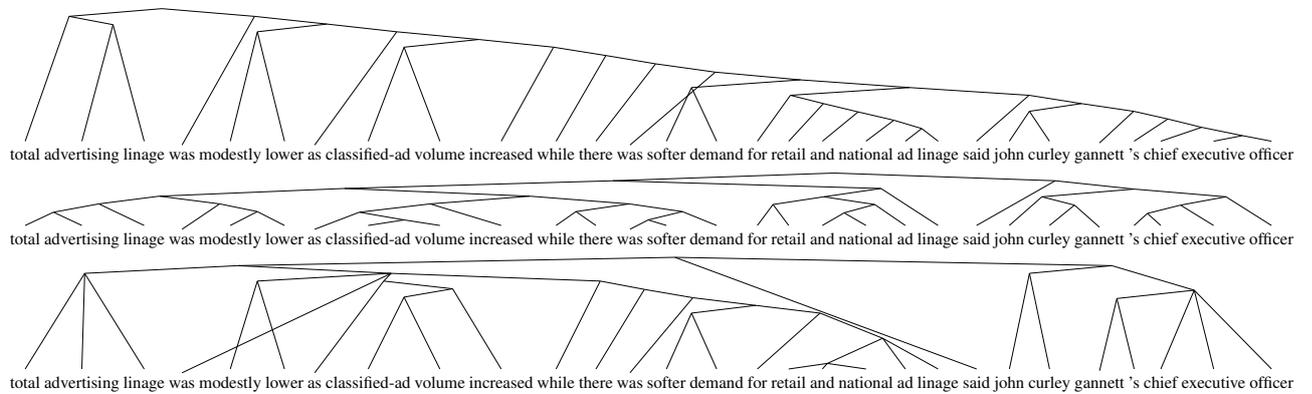

Figure 10: Sentence 6. Trees induced from the syntactic task (top) and language model task (middle) set of distances, as well as the gold-standard trees (bottom).

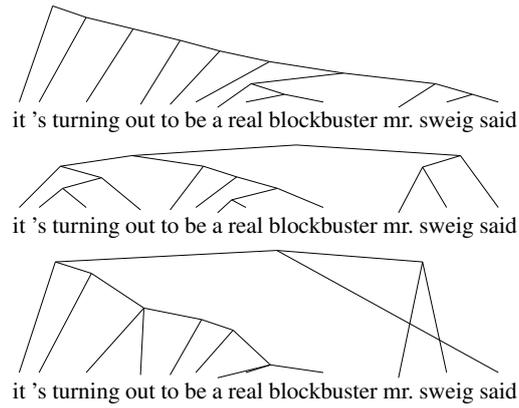

Figure 11: Sentence 7. Trees induced from the syntactic task (top) and language model task (middle) set of distances, as well as the gold-standard trees (bottom).

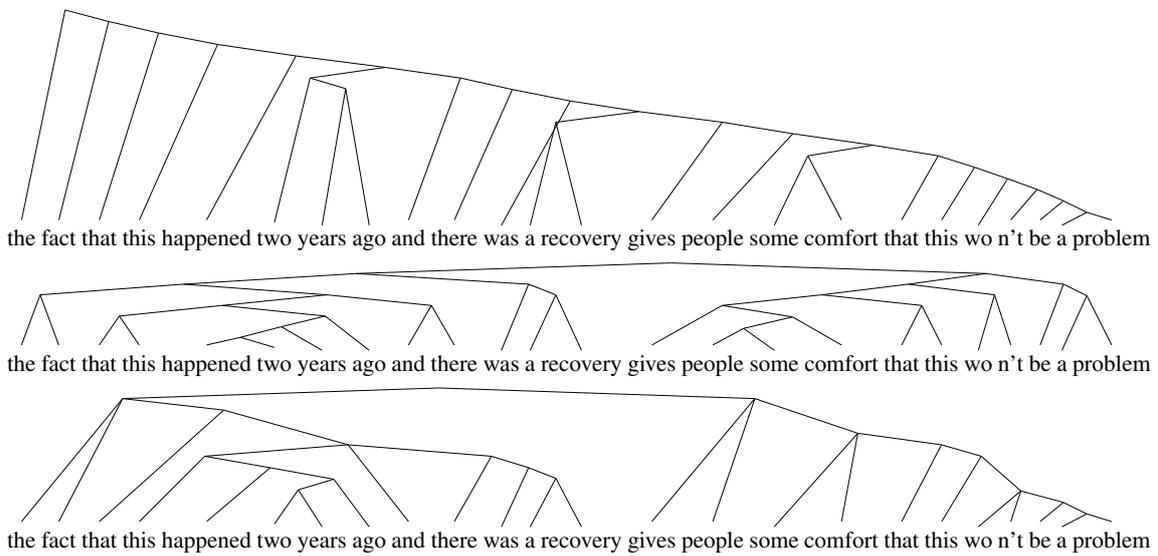

Figure 12: Sentence 8. Trees induced from the syntactic task (top) and language model task (middle) set of distances, as well as the gold-standard trees (bottom).

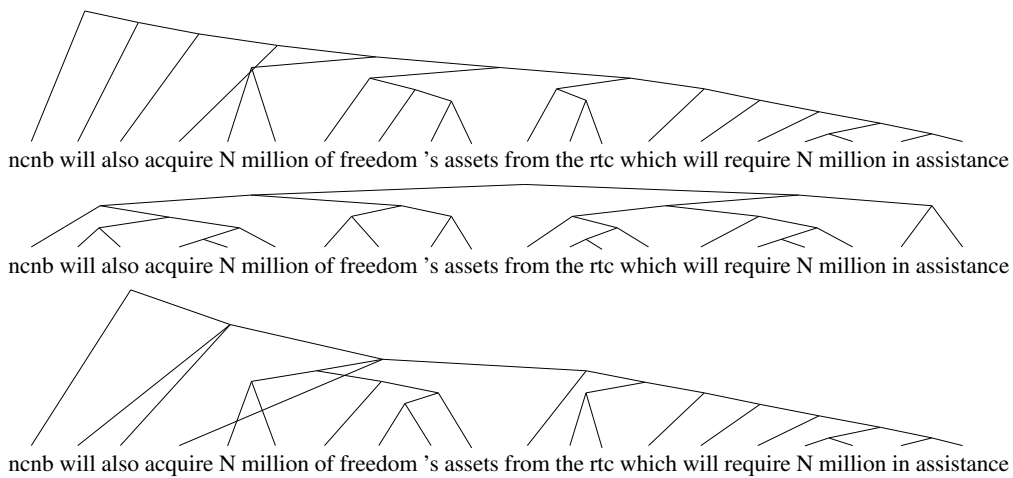

Figure 13: Sentence 9. Trees induced from the syntactic task (top) and language model task (middle) set of distances, as well as the gold-standard trees (bottom).

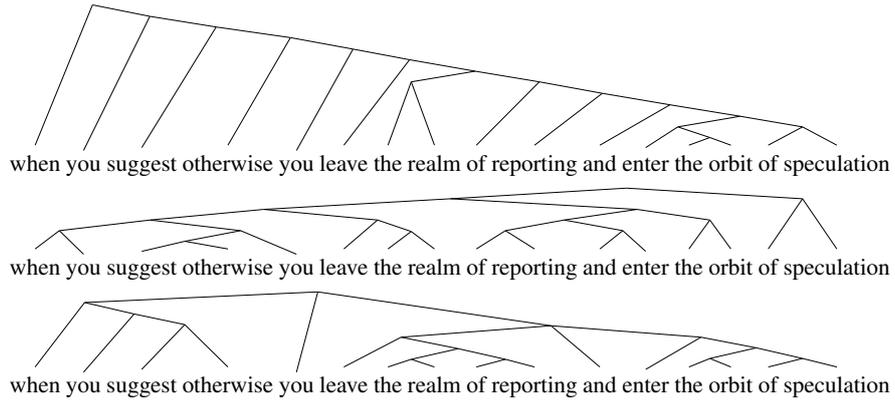

Figure 14: Sentence 10. Trees induced from the syntactic task (top) and language model task (middle) set of distances, as well as the gold-standard trees (bottom).

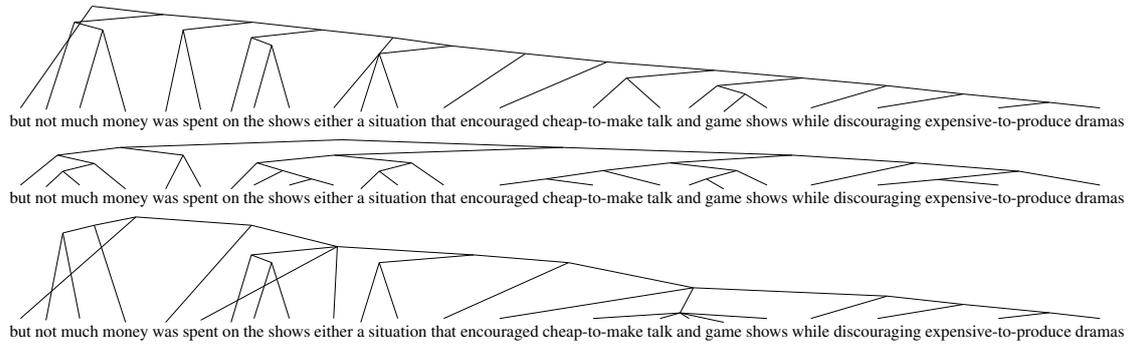

Figure 15: Sentence 11. Trees induced from the syntactic task (top) and language model task (middle) set of distances, as well as the gold-standard trees (bottom).

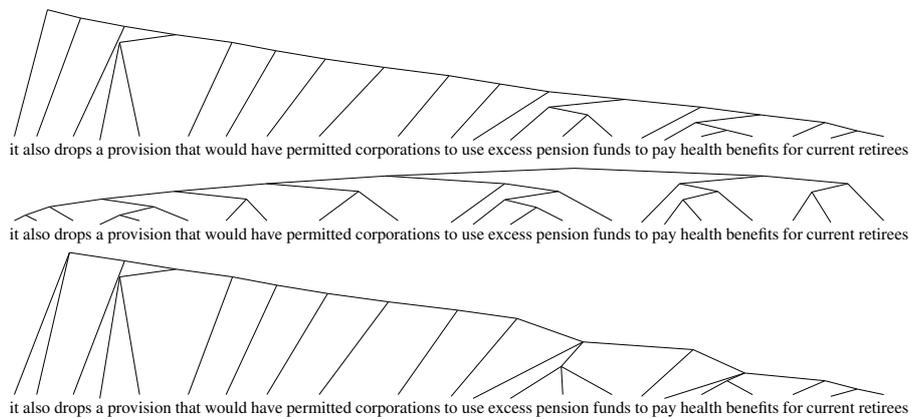

Figure 16: Sentence 12. Trees induced from the syntactic task (top) and language model task (middle) set of distances, as well as the gold-standard trees (bottom).

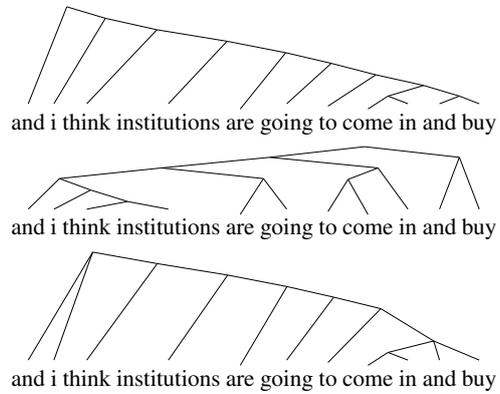

Figure 17: Sentence 13. Trees induced from the syntactic task (top) and language model task (middle) set of distances, as well as the gold-standard trees (bottom).

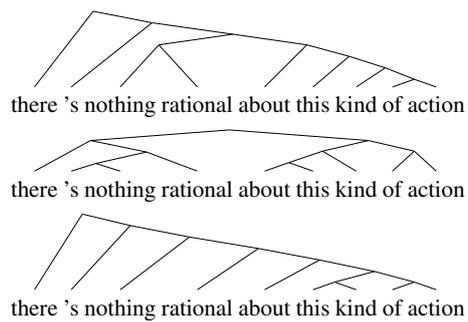

Figure 18: Sentence 14. Trees induced from the syntactic task (top) and language model task (middle) set of distances, as well as the gold-standard trees (bottom).